\documentclass[sigconf, table]{acmart}
\usepackage{bbm}
\usepackage{bm}
\usepackage{booktabs}
\usepackage{multirow}
\usepackage{caption}
\usepackage{subcaption}
\usepackage[export]{adjustbox}
\usepackage{xfrac}
\newcommand{\mat}{\bm}
\AtBeginDocument{%
  \providecommand\BibTeX{{%
    \normalfont B\kern-0.5em{\scshape i\kern-0.25em b}\kern-0.8em\TeX}}}

\setcopyright{acmcopyright}
\copyrightyear{2023} 
\acmYear{2023} 
\setcopyright{rightsretained} 
\acmConference[ICAIL 2023]{Nineteenth International Conference on Artificial Intelligence and Law}{June 19--23, 2023}{Braga, Portugal}
\acmBooktitle{Nineteenth International Conference on Artificial Intelligence and Law (ICAIL 2023), June 19--23, 2023, Braga, Portugal}\acmDOI{10.1145/3594536.3595131}
\acmISBN{979-8-4007-0197-9/23/06}

\begin{document}

\title{sustain.AI: a Recommender System to analyze Sustainability Reports}


\author{Lars Hillebrand}
\orcid{0000-0002-5496-4177}
\email{lars.patrick.hillebrand@iais.fraunhofer.de}
\author{Maren Pielka}
\author{David Leonhard}
\author{Tobias Deußer}
\affiliation{%
  \institution{Fraunhofer IAIS}
  \city{Bonn}
  \country{Germany}
}



\author{Tim Dilmaghani}
\author{Bernd Kliem}
\author{Rüdiger Loitz}
\affiliation{%
  \institution{PricewaterhouseCoopers GmbH}
  \city{Düsseldorf}
  \country{Germany}
}

\author{Milad Morad}
\author{Christian Temath}
\author{Thiago Bell}
\author{Robin Stenzel}
\author{Rafet Sifa}
\affiliation{%
  \institution{Fraunhofer IAIS}
  \city{Bonn}
  \country{Germany}
}

\renewcommand{\shortauthors}{Hillebrand and Pielka, et al.}  

\begin{abstract}
   We present \href{https://sustain.ki.nrw/}{sustain.AI}, an intelligent, context-aware recommender system that assists auditors and financial investors as well as the general public to efficiently analyze companies' sustainability reports. The tool leverages an end-to-end trainable architecture that couples a BERT-based encoding module with a multi-label classification head to match relevant text passages from sustainability reports to their respective law regulations from the Global Reporting Initiative (GRI) standards. We evaluate our model on two novel German sustainability reporting data sets and consistently achieve a significantly higher recommendation performance compared to multiple strong baselines. Furthermore, \href{https://sustain.ki.nrw/}{sustain.AI} is publicly available for everyone at \url{https://sustain.ki.nrw/}.
\end{abstract}





\maketitle

\section{Introduction}

In the face of climate change and environmental degradation, our society's expectations of sustainable and responsible entrepreneurial action have increased continuously over the past years. Legislators worldwide and particularly in the EU become increasingly aware of the situation and have taken concrete political measures to enforce corporate social responsibility (CSR). In 2014 the EU approved the Non-Financial Reporting Directive (NFRD) which forces large companies to extend their reporting on policies, risks and key performance indicators regarding sustainability and social matters. Beginning in 2024 the NFRD will be updated with the stricter CSR-Directive, which applies to around 50,000 European companies and includes a wider catalog of reporting requirements covering environmental, social and governance aspects.
The majority of these requirements are based on the popular regulatory framework from the Global Reporting Initiative (GRI). Its universal reporting standards provide a detailed set of indicators that address a company's impact on the economy, environment and people. 

In light of these more comprehensive and rigorous sustainability regulations and the public's growing interest in corporate social responsibility, it is of vital importance to make the disclosed information easily accessible and comparable. However, manually retrieving and analyzing the published reports concerning specific GRI-Indicators is practically infeasible, especially considering that the documents often span around a hundred pages or more. 
This is particularly true for the auditing domain, where auditors spend hours to assure a report's compliance related to said CSR standards.

\begin{figure}
\centering
  \includegraphics[width=\linewidth]{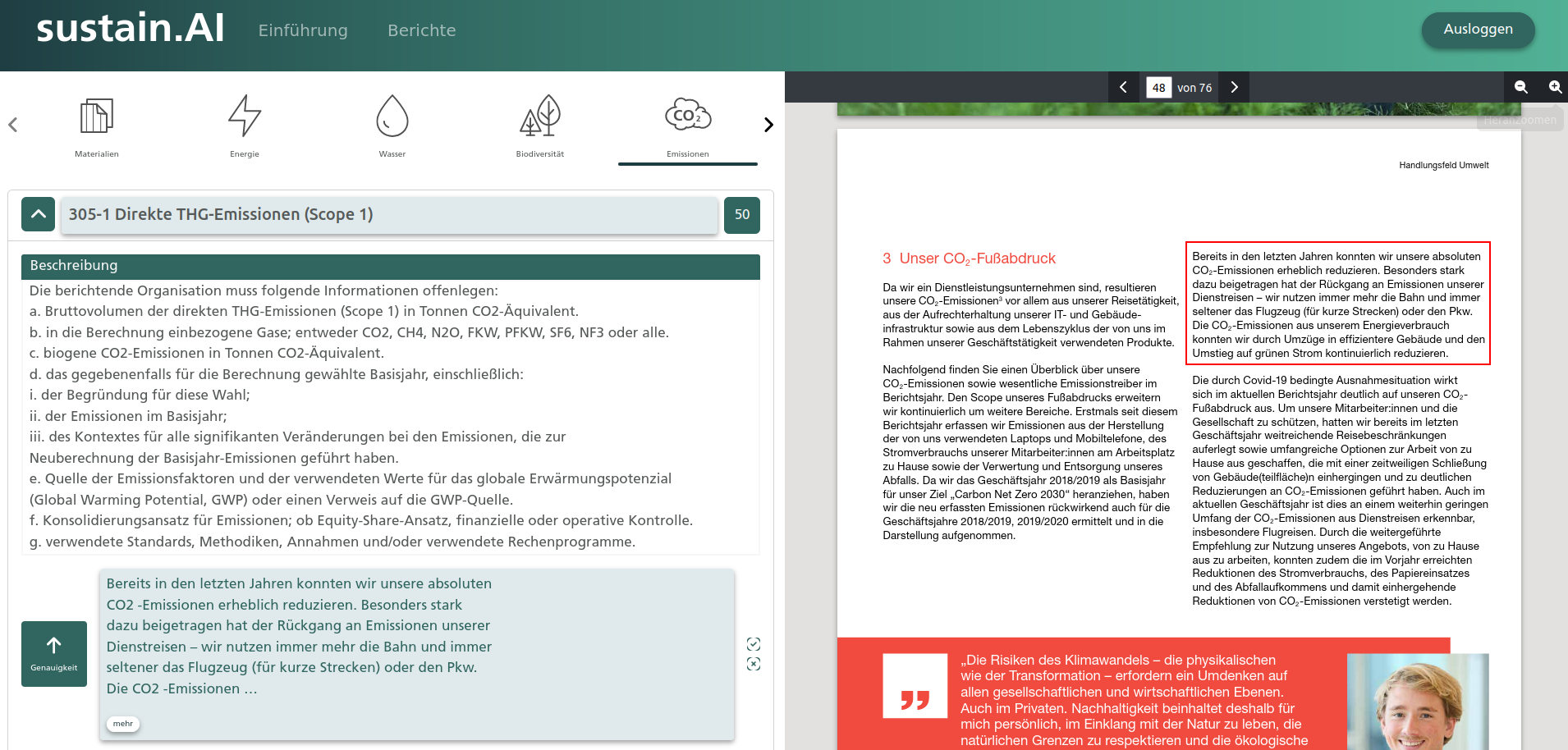}
  \caption{A screenshot of the sustain.AI recommender tool. After selecting a specific regulatory requirement from one of the categories, the system predicts the most relevant segments of a provided sustainability report. On the right side, the recommended segments are highlighted in the rendered report, fostering an efficient sustainability analysis.}
  \label{fig:sustain_screenshot}
\end{figure}

Hence, we introduce sustain.AI, a sophisticated, context-aware recommender system that utilizes modern techniques of natural language processing (NLP) and machine learning to process and analyze uploaded sustainability reports. Concretely, interested users like consumers or investors can query the recommender engine for specific GRI-indicators, e.g. the company's emissions (see Figure \ref{fig:sustain_screenshot}), and the engine returns and renders the most relevant document segments related to the query. Thus, stakeholders are able to quickly assess investment risks and opportunities arising from social and environmental issues and to evaluate the sustainability performance of companies. Similarly, auditors significantly benefit from the automated matching of concrete regulatory requirements to the relevant text passages. In fact, a large part of the sustainability report audit is about ensuring the completeness and correctness of the report according to the specified GRI standards.

Our recommender system builds on a BERT-based \cite{Devlin19} encoding module followed by a non-linear multi-label classification head. Both components are trained jointly in an end-to-end fashion leveraging weighted random sampling (WRS) to counter the significant class label imbalance. We evaluate the model on two novel German sustainability reporting data sets while consistently outperforming a large set of strong baselines by more than 10 percentage points in mean average precision.

sustain.AI is released to the public as a KI-NRW demonstrator, which is available at \url{https://sustain.ki.nrw/}. First user tests have already promised significant efficiency gains for the analysis of sustainability reports in the context of auditing. Moreover, the continuous use in production will further improve the system's recommendation capabilities due to the integration of human feedback, e.g. in the form of correcting wrong predictions.

\section{Related Work}

Before continuing with the description of the inner workings of sustain.AI, we take a look at prior accomplishments of other researchers related to this work.

In terms of facilitating the audit of annual financial statements, \cite{Sifa19} presented the Automated List Inspection (ALI) tool, a recommender system that ranks textual elements of financial documents to associated requirements of predefined regulatory frameworks like IFRS (International financial reporting standards) or HGB (Handelsgesetzbuch). For the ranking task, the authors used classical NLP techniques like Tf-Idf (Term frequency-Inverse document frequency), latent semantic indexing, neural networks and logistic regression (LR) with the combination of the first and last methods giving the best performance. In a follow-up work, \cite{Ramamurthy21} improved ALI by utilizing a pre-trained BERT \cite{Devlin19} language model as the backbone to encode text segments. Our architecture extends this approach by including weighted random sampling in the training process which speeds up the model convergence time and improves the overall performance. Concerning a more granular information extraction approach related to automatic consistency checks of financial disclosures, \cite{hillebrand2022towards} introduced KPI-Check, a BERT-based system that makes use of a tailored named entity and relation extraction model \cite{hillebrand2022kpi} to automatically detect and validate semantically equivalent key performance indicators in financial reports.

When it comes to the NLP-based analysis of sustainability or Corporate Social Responsibility (CSR) reports, different aspects have been researched. \cite{Gutierrez22} and \cite{Goel20} addressed the problem of automatically evaluating the GRI- and ESG\footnote{Environmental, Social and Governance factors.}-accordance of CSR-reports. Both applied unsupervised text similarity measures building on GloVe (Global Vectors for word representation) embeddings. Similarly, \cite{Angin22} leveraged the language model RoBERTa \cite{Liu19} to predict the relevance of sustainability reports according to the sustainable development goals in the USA. Specifically targeted for the banking sector, \cite{Moreno22} developed a rule-based named entity recognition approach to estimate an index that displays the level of compliance of the climate-related financial disclosures with the 
TCFD\footnote{Task Force on Climate-related Financial Disclosures.} recommendations.

\section{Methodology}

\begin{figure}[t]
  \centering
  \includegraphics[width=0.55\linewidth]{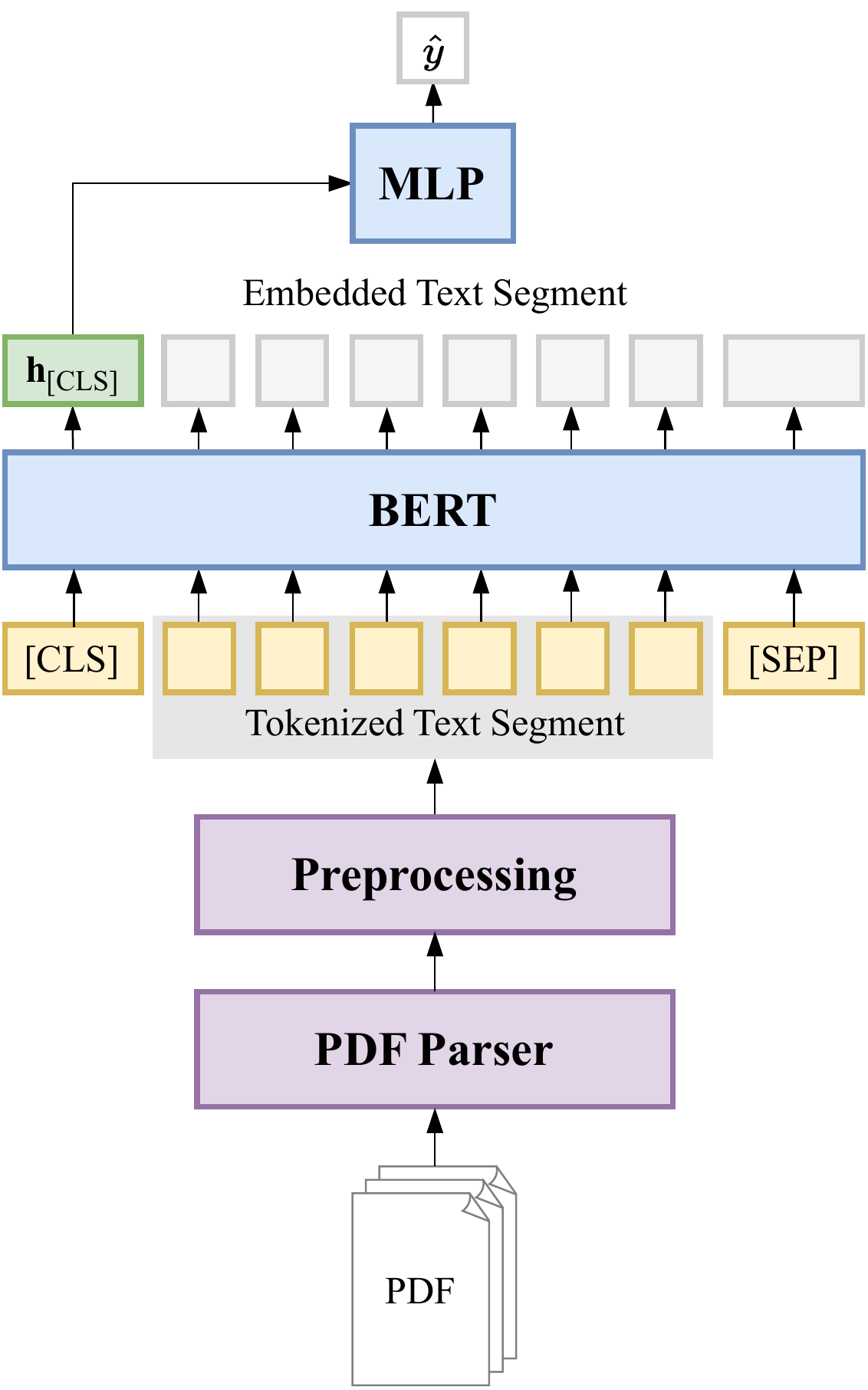}
  \caption{Schematical visualization of the recommender system and the data flow in sustain.AI. A custom PDF parser processes the raw sustainability reports. After some textual clean-ups, a fine-tuned BERT model encodes individual text segments that are subsequently matched to relevant regulatory requirements.}
  \label{fig:architecture}
\end{figure}

In this section, we formally define the problem of matching text segments within documents to relevant legal requirements before turning to the in-depth analysis of our proposed architecture which is visualized in Figure \ref{fig:architecture}.

\subsection{Problem Formulation}

Given a sustainability report consisting of $N$ distinct text segments $\mathcal{S}$, e.g. paragraphs, titles, tables or diagrams, and a set of $M$ regulatory checklist requirements $\mathcal{R}$, our goal is to identify all semantically relevant text segments for each requirement. Since the number of requirements $M$ is static, but each document has a different length (number of text segments) $N$, we initially model the described matching task from a segment-to-requirements perspective as a multi-label classification problem. Formally, for every $s_i \in \mathcal{S}$ our recommender model assigns relevance scores to all $r_j \in \mathcal{R}$. 

However, from the users' point of view, the reverse direction of getting relevant segment recommendations for a specific requirement $r_j$ (requirement-to-segments perspective) is far more beneficial. This is especially true because a significant amount of text segments within a sustainability report is unrelated to concrete requirements in $\mathcal{R}$. This is why, based on the assigned relevance scores, our model ranks the text segments per requirement in descending order and subsequently recommends the top $K$ relevant text blocks to the user. 

\subsection{Document Parsing}

Before we focus on the actual recommender module, the core component of sustain.AI, we briefly touch upon the non-negligible task of document parsing. The large majority of publicly available sustainability reports are published as PDF documents, an inherently difficult format to convert into a structured machine-readable form like XML or JSON. The latter is particularly true for scanned PDF reports that only contain image information. 

To solve this issue our system utilizes a custom PDF parser (see Figure \ref{fig:architecture}), that is capable of parsing machine-created as well as scanned PDFs with arbitrarily complex formattings.
The parser leverages a refined image segmentation technique by combining the powerful object detection network Faster R-CNN \cite{ren2015faster} with the density-based clustering algorithm DBSCAN \cite{ester1996density}. It is also trained to recognize specific elements of a document, such as footers, headers or pagination. For further details about the parser's functionality, we refer to \cite{agombar2020clustering}.

After the successful PDF parsing we apply some basic textual preprocessing in the form of removing line break hyphens and filtering out irrelevant text segment types like footer, header and table of contents. Our final set of considered segments $\mathcal{S}$ consists of titles, paragraphs, enumerations, tables and diagrams.

\subsection{Recommender System}
\label{sec:model}

Considering a parsed and processed sustainability report, we use a pretrained BERT \cite{Devlin19} model to individually encode each text segment $s_i \in \mathcal{S}$. 

Formally, we first apply WordPiece \cite{schuster2012japanese} tokenization to transform an examplary input segment $s$ into a sequence of sub-word tokens $t=\left(\text{[CLS]}, t_1, \dots, t_n, \text{[SEP]}\right)$. Note [CLS] denotes a BERT-specific special token that aggregates the content of the entire segment while [SEP] simply highlights the end of the sequence.

Passing $t$ to the BERT model with pretrained parameters $\mat{W}_{\text{bert}}$ yields a sequence of contextual token embeddings $\mat{h}_{[\text{CLS}]}, \mat{h}_1, \dots, \allowbreak \mat{h}_n, \mat{h}_{[\text{SEP}]}$, where $\mat{h}_{[\text{CLS}]}$ represents the aggregated context hidden state for the whole segment $s$. 

Subsequently, we employ a multi-layer perceptron (MLP) with trainable parameters $\mat{W}_{\text{mlp}}$ to predict relevance probabilities $\hat{\mat{y}}=[\hat{y}_1, \dots, \hat{y}_M] \in \mathbb{R}^M$ for all requirements in $\mathcal{R}$.
The classifying MLP consists of a fully-connected hidden layer followed by dropout and ReLU (Rectified Linear Unit) activation functions and a sigmoidal output layer.

During training, we jointly optimize and finetune the parameters of the BERT model $\mat{W}_{\text{bert}}$ and the classification layer $\mat{W}_{\text{mlp}}$ to minimize the Binary Cross Entropy (BCE) loss between target labels $\mat{y}$ and predicted probabilities $\hat{\mat{y}}$.

Finally, after assigning relevance scores over requirements for all $s_i \in \mathcal{S}$, we sort the segments for each requirement $r_j$ in descending order in order to recommend the top $K$ relevant text blocks.

\section{Experiments}

In the following sections, we introduce our two custom data sets of German sustainability reports, define our evaluation metrics, discuss the overall training setup, describe the competing baseline methods, and finally, evaluate results.

\subsection{Data}

We train and evaluate our algorithms on two novel sustainability reporting data sets. 

The first data set, named GRI, consists of 92 published sustainability reports from major German companies. The reports have been sourced in PDF format from the companies' websites. After the parsing step domain experts from the auditing industry annotated all text segments in accordance to the requirements of the Global Reporting Initiative (GRI) standards. Concretely, we consider the 89 indicators of the GRI topic standards which cover the three main categories, economy, environment and social that are further split into granular topics like anti-corruption, energy consumption and human rights assessment. The annotation work load was equally split among three auditors which were supervised by a senior auditor. In multiple iterations, the created requirement labels have been validated and refined via double-checking randomly selected sample annotations and a qualitative inspection of the false positive and negative model predictions.

The second data set, named DNK, leverages the public sustainability reporting database from the German Sustainability Code\footnote{\url{https://www.deutscher-nachhaltigkeitskodex.de/Home/Database}.} (DNK). The platform is used by the majority of German companies to annually disclose their sustainability activities with respect to 33 requirements from 20 DNK criteria, e.g. usage of natural resources and human rights. The categories and their requirements cover most of the GRI topics but are generally less granular. 
In contrast to the PDF documents of the GRI data set, the DNK reports in HTML format follow a predefined structure where each section of text segments answers a distinct requirement. Since the requirement descriptions precede their respective sections we can automatically retrieve the ground truth annotations from the HTML during the parsing process.

\begin{table}[t]
\small
    \caption{Properties of our GRI and DNK data sets. We display the number of requirements and documents, the average number of segments per document, the average percentage of segments assigned to at least one requirement, and the average number of matched segments per requirement.}
    \centering
    \label{tab:dataset_properties}
    \begin{tabular}{lcc}
        \toprule
        Data set & GRI & DNK\\
        \toprule
        \# requirements & 89 & 33\\
        \# documents & 92 & 1779\\
        \# segments $s$ per document & 972 & 242 \\
        \% segments $s$ matched & 9 & 100\\
        \# matched segments $s$ per requirement & 2.7 & 7.3 \\
        \bottomrule
\end{tabular}
\end{table}

Table \ref{tab:dataset_properties} displays descriptive statistics for both data sets. 
Due to the smaller amount of training documents, the greater document size and the annotation sparsity, we consider the GRI data set the harder challenge for our models. 
We separately train, optimize and evaluate our algorithms on both data sets to verify this hypothesis, investigating how well sustain.AI handles different sizes of training data and number of labels. For our GRI and DNK experiments, we employ fixed training, validation and testing splits of 65-15-20 and 70-15-15, respectively.

As a contribution to the open-source community and for further research concerning German sustainability reports we make the DNK data set publicly available\footnote{\url{https://github.com/LarsHill/dnk-dataset}.}.

\subsection{Evaluation Metrics}

We quantitatively evaluate all models by calculating modified mean sensitivity (MS) and mean average precision (MAP) scores for the top $K$ recommendations. While MAP punishes the lower ranked recommendations of relevant segments, MS only considers whether the relevant segments are contained in the set of recommendations. For a single document and a concrete requirement $r_j$ the modified sensitivity S($K$) from \cite{Sifa19} and the average precision AP($K$) are respectively defined as:
\begin{align}
    \text{S}(K) &= \frac{\lvert \text{top } K \text{ recommendations} \cap L \text{ annotations} \rvert}{\text{min}(K, L)},  \\
    \text{AP}(K) &= \frac{1}{\text{min}(K, L)}\sum_{i=1}^K \left( \text{P}(i) \cdot \text{rel}(i)\right),
\end{align}
where $L$ denotes the number of relevant segment annotations, rel($i$) indicates whether the $i^{\text{th}}$ recommendation is relevant (rel$(i)=1$) or not (rel$(i)=0$), and 
\begin{align}
    \text{P}(i) = \frac{\lvert \text{top } i \text{ recommendations} \cap L \text{ annotations} \rvert}{i}
\end{align}
represents the precision score considering the top $i$ recommendations. Averaging S$(K)$ and AP$(K)$ over all checklist requirements $r_j \in \mathcal{R}$ and documents yields the subsequently reported mean sensitivity MS$(K)$ and mean average precision MAP$(K)$ metrics.

\subsection{Training Setup}

In this section, we shed light on the training process and the hyperparameter optimization of sustain.AI.

For all evaluated models we conduct an exhaustive grid search comparing various parameter combinations based on their validation set MAP$(3)$ performance to determine the best training setup. Table \ref{tab:hyperparameters} highlights the explored ranges and respective best values of sustain.AI's tuned model parameters.
\begin{table}[t]
\small
    \caption{Evaluated hyperparameter configurations of sustain.AI. The best configuration on the validation set is highlighted in boldface.}
    \centering
    \label{tab:hyperparameters}
\begin{tabular}{lc}
\toprule
Hyperparameter &  Configurations \\
\midrule
MLP hidden dimensions & None, $512$, $\bm{1024}$, $2048$ \\
Dropout         & $0.0$, $0.1$, $\bm{0.3}$, $0.5$, \\
Batch size      & $2$, $4$, $\bm{8}$, $16$ \\
Learning rate   & $1e{-4}$, $\bm{1e{-5}}$, $1e{-6}$ \\
\bottomrule
\end{tabular}
\end{table}

As encoding backbone we employ a BERT$_{\text{BASE}}$ model,
published by the MDZ Digital Library team (dbmdz)\footnote{ \url{https://huggingface.co/dbmdz/bert-base-german-cased}.}. It mirrors the architectural setup of the English BERT$_{\text{BASE}}$ counterpart\footnote{12 multi-head attention layers with 12 attention heads per layer and 768-dimensional output embeddings.} and is pre-trained on a large corpus of German books, news reports and Wikipedia articles. We train our model and all neural network based baselines via gradient descent utilizing the AdamW \cite{loshchilov2017decoupled} optimizer with a linear warmup of $10$\% and a linearly decaying learning rate schedule. Additionally, we apply weight decay of 0.01 and gradient clipping with a maximum value of 1. We also analyze different learning rates, batch sizes, levels of dropout regularization, and MLP hidden dimensions, as can be seen in Table \ref{tab:hyperparameters}. For all training runs we set a random seed of 42 and fix the maximum number of epochs to 15 while applying early stopping with a patience of 3 epochs.

Due to the small percentage of annotated segments $s$ in the GRI data set (9\%, see Table \ref{tab:dataset_properties}) we employ weighted random sampling (WRS) with replacement to expose these relevant segments more frequently during training. Concretely, we alter the originally uniform sampling probability of each segment to the normalized inverse frequency of relevant $+$ or irrelevant $-$ occurrences in the training set. 
\begin{figure}[t]
\centering
\begin{subfigure}[t]{.5\linewidth}
  \centering
  \includegraphics[width=0.95\linewidth, left]{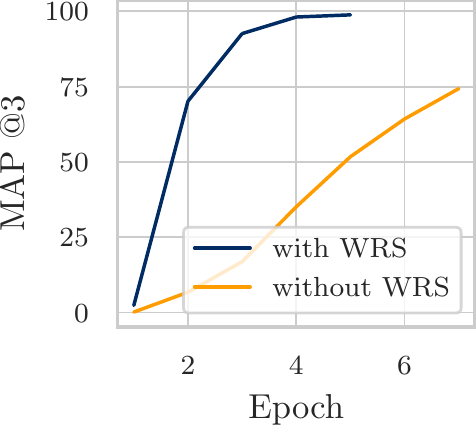}
  \caption{Training}
  \label{fig:sub1}
\end{subfigure}%
\begin{subfigure}[t]{.5\linewidth}
  \centering
  \includegraphics[width=0.95\linewidth, right]{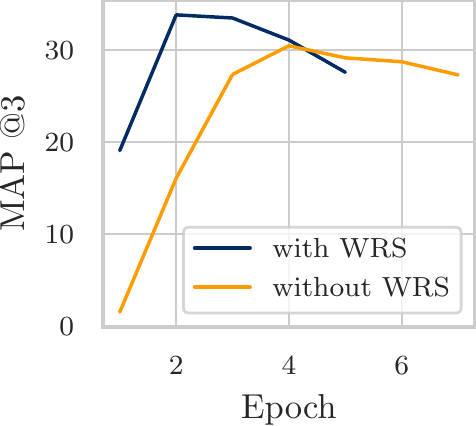}
  \caption{Validation}
  \label{fig:sub2}
\end{subfigure}
\caption{Positive impact of weighted random sampling (WRS) on training convergence and validation performance. We report the mean average precision considering the top 3 recommendations (MAP$(3)$) with and without WRS.}
\label{fig:wrs_improvement}
\end{figure}

Figure \ref{fig:wrs_improvement} showcases the benefits of integrating WRS into the model training process for the GRI data set. We achieve a much faster training convergence and thus, save a considerable amount of training time and compute power benefitting from early stopping. 
At the same time, our model's MAP$(3)$ score on the validation set increases by 3 percentage points.

\subsection{Baselines}

We compare sustain.AI's end-to-end recommender model from Section \ref{sec:model} with 4 competing baseline architectures. 
For a fair comparison, all baselines make use of weighted random sampling concerning the imbalanced GRI data set.

First, we utilize word frequency-based Tf-Idf \cite{ramos2003using} 
representations that have been fitted on our respective training corpora. Prior to training, all segments have been preprocessed in terms of lowercasing, punctuation- and digit removal as well as stemming. The resulting 8000 dimensional segment vectors are then used as input for an ensemble of one-vs-rest binary logistic regression (LR) classifiers. Each classifier is trained for a specific requirement $r$ and a maximum of 100 iterations using the ``liblinear'' solver from the scikit-learn python library.

Second, we pass the same Tf-Idf representations into an MLP with one hidden layer of dimensionality 1024. In contrast to the binary logistic regression heads, the MLP performs multi-label classification and predicts the relevant requirements simultaneously. 
We find an optimal batch size of 64 and a learning rate of $1e{-3}$.

Third, we exchange the Tf-Idf input vectors with frozen contextual embeddings from sustain.AI's BERT model. As classifiers we evaluate the previously defined MLP and a GRU (Gated Recurrent Unit). While the MLP takes BERT's CLS output embedding as input, the bidirectional GRU processes the resulting token representations of the frozen BERT model. Specifically, the last/first hidden state of the forward/backward GRU are concatenated and passed to a sigmoidal output layer. Optimal settings are obtained with a hidden size of 512 neurons, a batch size of 8 and a learning rate of $1e{-5}$.

\subsection{Results}

\begin{table}[t]
\small
\centering
\caption{Test set results for the recommendation of relevant segments in GRI and DNK sustainability reports. sustain.AI outperforms all competing baselines in top 3/5 mean sensitivity (MS) and mean average precision (MAP). 
}
\label{tab:model_results}
\begin{tabular}{lc@{\hspace{0.55em}}cc@{\hspace{0.55em}}cc@{\hspace{0.55em}}cc@{\hspace{0.55em}}c}
\toprule
in \% & \multicolumn{4}{c}{GRI} & \multicolumn{4}{c}{DNK} \\ 
\cmidrule(lr){2-5}
\cmidrule(lr){6-9}
\multirow{2}{*}{Model} & \multicolumn{2}{c}{MS} & \multicolumn{2}{c}{MAP} & \multicolumn{2}{c}{MS} & \multicolumn{2}{c}{MAP} \\
\cmidrule(lr){2-3}
\cmidrule(lr){4-5}
\cmidrule(lr){6-7}
\cmidrule(lr){8-9}
           & 3 & 5 & 3 & 5 & 3 & 5 & 3 & 5 \\
\midrule
Tf-Idf + LR          & 24.3 & 33.7 & 17.1 & 19.5 & 70.8 & 66.3 & 66.8 & 59.0 \\
Tf-Idf + MLP         & 33.0 & 39.8 & 22.6 & 24.2 & 77.4 & 77.8 & 74.8 & 71.8 \\
BERT$_{\text{frozen}}$ + MLP  & 28.4 & 36.1 & 21.0 & 22.4 & 75.2 & 70.6 & 73.5 & 66.4 \\
BERT$_{\text{frozen}}$ + GRU  & 28.1 & 36.8 & 20.5 & 22.1 & 84.0 & 80.2 & 83.0 & 77.2 \\
sustain.AI$_{\text{no WRS}}$  & 35.5 & 44.2 & 28.4 & 30.5 & $\bm{90.3}$ & $\bm{87.8}$ & $\bm{89.7}$ & $\bm{86.1}$ \\
sustain.AI$_{\text{WRS}}$     & $\bm{48.0}$ & $\bm{53.8}$ & $\bm{35.9}$ & $\bm{37.0}$ & - & - & - & - \\
\bottomrule
\multicolumn{9}{l}{\scriptsize{{\small-} $=$ not applicable, since weighted random sampling (WRS) is only applied on GRI data.}}
\end{tabular}

\end{table}
We evaluate and compare sustain.AI and all baseline methods on the previously specified hold out test set for both the GRI and DNK data. Table \ref{tab:model_results} reports mean sensitivity (MS) and mean average precision (MAP) scores for the top 3 and top 5 recommendations.

First, it can be seen that the overall DNK performance across all methods is much better compared to 
the GRI data. This was expected, considering the reduced number of requirements and the larger amount of training documents and annotations.

Second, we find that the application of weighted random sampling (WRS) during training significantly improves the test set performance of our model. Compared to the version without WRS all metrics have increased by more than 6 percentage points. To enable a fair comparison we apply WRS during the training process of all baseline methods. Also, WRS is solely employed for the GRI data, since the DNK reports do not exhibit any annotation scarcity.

Finally, the results in Table \ref{tab:model_results} show the overall superiority of sustain.AI's end-to-end architecture, outperforming all baselines by a large margin.

\section{Conclusion and Future Work}
We presented sustain.AI, an interactive, AI-powered tool for the semi-automated analysis of German sustainability reports. Our transformer-based model achieves promising results both on the well-structured DNK data set and on the real-world GRI data, compared to a number of strong baselines. Qualitative exploration of the results also suggests that it is indeed helpful in analyzing those long documents. The tool is planned to be deployed on an online platform soon and will then be openly accessible to the public.

Future work includes improving the current model with additional annotated data, which can easily be inferred from the user feedback we will collect through the tool. We also plan to extend the framework to English reports, as currently only the processing of German documents is possible. Another idea for improvement is to extract specific numeric key performance indicators from the reports, such as different types of CO$_2$ emissions, water consumption or indicators for social welfare.

\begin{acks}
This research has been partially funded by the Federal Ministry of Education and Research of Germany and the state of North-Rhine Westphalia as part of the Lamarr-Institute for Machine Learning and Artificial Intelligence.
\end{acks}

\bibliographystyle{ACM-Reference-Format}
\bibliography{bibliography}  




\end{document}